\documentclass[letterpaper, 10 pt, conference]{ieeeconf}  

\IEEEoverridecommandlockouts                              

\usepackage{graphics} 
\usepackage{epsfig} 
\usepackage{mathptmx} 
\usepackage{times} 
\usepackage{amsmath} 
\usepackage{amssymb}  
\usepackage{color}
\usepackage{lipsum}  
\usepackage{relsize}
\usepackage{url}
\usepackage{siunitx}

\usepackage{bbm}
\usepackage{algorithm}
\usepackage{algorithmic}
\usepackage{newtxtext, newtxmath}
\usepackage{float}

\DeclareMathOperator*{\argmin}{arg\,min}

\usepackage{hyperref}
\hypersetup{
    colorlinks=true,
    linkcolor=blue,
    filecolor=magenta,      
    urlcolor=cyan,
}

\title{\LARGE \bf
Scalable Multi-Task Imitation Learning with Autonomous Improvement
}

\author{Avi Singh*$^{1}$, Eric Jang$^{2}$, Alexander Irpan$^{2}$, Daniel Kappler$^{3}$, Murtaza Dalal$^{1}$,\\ Sergey Levine$^{1,2}$, Mohi Khansari$^{3}$, Chelsea Finn$^{2,4}$
\thanks{*Work done while an AI Resident at X, Mountain View.}
\thanks{$^{1}$UC Berkeley, $^{2}$Robotics at Google, $^{3}$X$, ^{4}$Stanford}%
}

\begin{document}

\maketitle
\thispagestyle{empty}
\pagestyle{empty}


\begin{abstract}
While robot learning has demonstrated promising results for enabling robots to automatically acquire new skills, a critical challenge in deploying learning-based systems is scale: acquiring enough data for the robot to effectively generalize broadly. Imitation learning, in particular, has remained a stable and powerful approach for robot learning, but critically relies on expert operators for data collection.
In this work, we target this challenge, aiming to build an imitation learning system that can continuously improve through autonomous data collection, while simultaneously avoiding the explicit use of reinforcement learning, to maintain the stability, simplicity, and scalability of supervised imitation.
To accomplish this, we cast the problem of imitation with autonomous improvement into a multi-task setting. We utilize the insight that, in a multi-task setting, a failed attempt at one task might represent a successful attempt at another task. This allows us to leverage the robot's own trials as demonstrations for tasks other than the one that the robot actually attempted.
Using an initial dataset of multi-task demonstration data, the robot autonomously collects trials which are only sparsely labeled with a binary indication of whether the trial accomplished any useful task or not.
We then embed the trials into a learned latent space of tasks, trained using only the initial demonstration dataset, to draw similarities between various trials, enabling the robot to achieve one-shot generalization to new tasks. 
In contrast to prior imitation learning approaches, our method can autonomously collect data with sparse supervision for continuous improvement, and in contrast to reinforcement learning algorithms, our method can effectively improve from sparse, task-agnostic reward signals.
\end{abstract}
\section{Introduction \& Related Work}

\begin{figure}[t]
\centering
  \includegraphics[width=1.0\columnwidth]{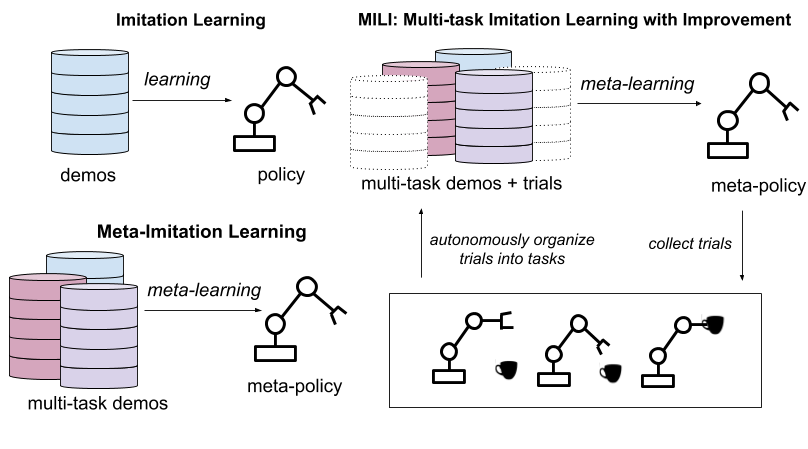}
  \vspace{-0.8cm}
  \caption{
  In this figure, we describe our problem setting and differentiate it from the standard imitation learning and meta-imitation learning problem settings.
  In standard imitation learning, we are provided a dataset of demonstrations for a specific task, and we learn a policy to mimic the behavior in the demonstrations. In meta-imitation learning, we are provided with several such datasets drawn from a task distribution, and we learn a single meta-policy that is capable of performing can generalize to new tasks in this distribution from a single demonstration. We also consider a one-shot imitation learning setting in this paper; however, unlike standard meta-imitation, our method can also utilize autonomously collected experience. After filtering out trials that do not succeed on any task, our method adds the autonomously collected trials to the meta-training set, matching them with existing tasks using an automated pairing method. We use this newly collected and organized data to update the policy, establishing a learning loop that can perpetually improve from autonomously collected data.}
  \vspace{-0.6cm}
  \label{fig:problemsetting}
\end{figure}

Robotic learning holds the potential to enable \emph{generalist} robots: robotic systems that can autonomously perform a wide range of different behaviors. 
In order to enable robots to be generalists, equipped with large repertoires of skills, we need them to be able to acquire each skill from a relatively modest amount of experience, and to make it feasible to add new skills using only a little bit of data.
Such generalist systems not only have tremendous promise for applications across a range of domains that cannot currently be automated, but would also represent a significant step forward in artificial intelligence research. Two of the most successful approaches to general-purpose robotic learning are imitation learning, where a robot uses human-provided demonstration data to learn skills~\cite{pomerleau1989alvinn, vr_imitation}, and reinforcement learning (RL), where it uses a trial-and-error learning process~\cite{sutton1998reinforcement,kober2012reinforcement}.

When combined with high-capacity function approximators, such as deep neural networks, both techniques have been shown to enable complex robotic skills directly from low-level sensory observations, such as images~\cite{vr_imitation, kalashnikov2018qt}. However, both techniques also have substantial limitations, some of which we will aim to mitigate in this work. First, the combination of deep networks with imitation learning and reinforcement learning results in very high data complexity. In imitation learning, this means that human operators must provide a large number of demonstrations for each task, while in the case of reinforcement learning, this translates into very large autonomous data collection requirements particularly to overcome the challenges of exploration~\cite{mnih2013playing}.
With the data requirements in both cases, it is thus difficult to scale these techniques to enable truly generalist robots equipped with large repertoires of skills. The two approaches also have some distinct strengths and weaknesses. Reinforcement learning enables a robot to improve, perhaps perpetually, from its own experience~\cite{kalashnikov2018qt, haarnoja2018sacapps, singh2019}, but RL-based policies are generally more limited in the complexity of the tasks they can accomplish, due to the difficulty of discovering task solutions autonomously and the complexity of the RL optimization problem~\cite{henderson2018deep,irpan2018rlblogpost,dulac2019challenges}. By comparison, imitation learning methods present a much simpler \emph{supervised} learning problem, making them suitable for more complex skills~\cite{sun2017deeply, vr_imitation}, but they require a more manual process of collecting human demonstration data~\cite{vr_imitation},
and the robot cannot continue to improve autonomously from its own experience~\cite{ross2010efficient}. This last point in particular is a major shortcoming of current imitation learning approaches: the capacity for continuous and autonomous self-improvement is an exceptionally powerful feature of RL.

\begin{figure}[t]
\centering
  \includegraphics[width=0.5\textwidth]{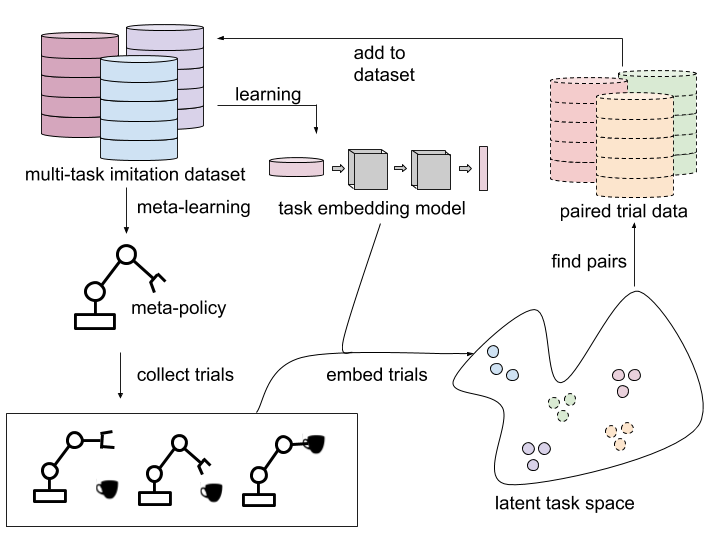}
  \vspace{-0.8cm}
  \caption{The MILI algorithm. We bootstrap a one-shot imitation policy using a multi-task imitation learning dataset. We then use this policy to collect trials in new environments. A latent task space, learned using the same initial dataset, is then used to find similarities in the collected trials, and generate new tasks for meta-imitation learning. We update our meta-policy using the newly collected data, and repeat this process until convergence.}
  \vspace{-0.6cm}
  \label{fig:pipeline}
\end{figure}

Multi-task~\cite{multitaskpolicy} and meta-imitation learning~\cite{rocky2017, finn2017one, yu2018one, james2018task, taskgraph18,wtl,paine2018one} alleviate the data collection problem to some extent by reusing data across many different tasks. These methods typically learn a parameterized policy that is a function of both the current observation and the task, thus obtaining a single policy that is capable of performing several different tasks and generalizing to new tasks. For multi-task imitation learning, the task is typically specified in the form of a task index, while in meta-imitation, the task is specified using a single demonstration for that task. Thus, meta-imitation learning systems learn a one-shot imitation policy, capable of performing a new task from just a single demonstration for it. However, these approaches still require manually collecting data for hundreds of different tasks, and cannot improve from autonomously collected data. In this paper, we will also consider a one-shot imitation learning problem. However, unlike these prior works, our goal is to develop an algorithm that can improve its one-shot imitation performance using additional \emph{autonomously-collected} data.

To this end, we utilize the following insight: if a robot is trained in a multi-task setting consisting of multiple distinct tasks with hundreds of different objects, it is likely to perform some useful behavior when it attempts a task in a new setting, even if it is not the behavior that was actually intended. This data can then be leveraged to learn the skill that was actually performed. This insight can be viewed as a generalization of prior works on goal-conditioned learning via hindsight goal relabeling~\cite{kaelbling1993goals,andrychowicz2017hindsight,yu2019unsupervised,rauber2017hindsight},
but in the space of \emph{tasks} rather than the space of goals to be reached. It is this insight, and how we can leverage it in a meta-imitation learning framework, that provides the novel contribution of this work: a framework for autonomous data collection and improvement \emph{without} explicit reinforcement learning, that instead leverages a supervised meta-imitation learning method to enable a robot to be bootstrapped off of demonstration data, and then improve its repertoire of skills via its own autonomously collected experience.

Our approach, which we call Multi-task Imitation Learning with Improvement (MILI), is summarized in Figure~\ref{fig:pipeline}. We bootstrap a meta imitation policy by performing one-shot imitation learning on a human-collected dataset of paired demonstrations, which consists of pairs of optimal trajectories corresponding to the same task (following the standard meta-imitation learning set-up in prior work~\cite{duan2017one,finn2017one}).
We then use this policy to collect trials autonomously. For each such trial, a user needs to provide an answer to the following yes/no question: did the robot succeed at \textit{any} task of interest? If the answer is yes, this trajectory is added to an augmented meta-training set. This binary label can be provided via an automated reward function or direct human feedback. We use the initial demonstration dataset of paired demonstrations to also learn a latent space of tasks, where demonstrations corresponding to the same tasks are pushed close to each other, and demonstrations corresponding to different tasks are pushed apart. We then embed all of the autonomously collected trajectories into this learned task space, and generate a dataset of paired trajectories by pairing trajectories that are close to each other in the latent space. 

The core contribution of this paper is a framework that allows imitation learning systems to continuously improve through autonomous data collection and learning.
We empirically study a challenging, vision-based manipulation setting consisting of hundreds of tasks from four distinct task families: button pressing, sliding, grasping, and pick-and-place.  Our framework allows for substantial improvements over standard imitation learning and meta-imitation learning approaches and, critically, has the potential to continuously improve itself without human-provided demonstrations.
\newcommand{\data}{\mathcal{D}}
\newcommand{\demo}{d}
\newcommand{\traj}{\tau}
\newcommand{\metatest}{\mathcal{D}_{\mathcal{T}_\text{test}}}
\newcommand{\metatrain}{\mathcal{D}_{\mathcal{T}_\text{train}}}
\newcommand{\task}{\mathbb{T}}
\newcommand{\loss}{\mathcal{L}}
\newcommand{\bcloss}{\loss_\text{bc}}
\newcommand{\norm}[1]{\left\lVert#1\right\rVert}
\newcommand{\obs}{o}
\newcommand{\act}{a}
\newcommand{\fulldataset}{\left\{\data_i\right\}}

\section{Preliminaries}
\label{sec:prelim}

In this section, we summarize the one-shot imitation learning problem setting considered in prior work~\cite{rocky2017, finn2017one, paine2018one, taskgraph18, james2018task}. We first formalize the single-task imitation learning setting, and then extend it to the one-shot setting.

\noindent \textbf{Imitation Learning via Behavior Cloning.}
In the standard single task imitation learning setting~\cite{pomerleau1989alvinn,imitation_survey}, we are provided a dataset of expert demonstrations $\data = \left\{d_1, d_2,...d_N \right\}$. Each demonstration $\demo_n$ consists of a trajectory of observations and actions denoting optimal behavior for that task $\demo_n = \left[\left(\obs_1^n, \act_1^n\right), \cdots, \left(\obs_T^n, \act_T^n\right)\right]$, and we need to learn a policy $\pi$ that can mimic this expert behavior. While there are several ways to perform imitation learning from expert demonstrations such as inverse reinforcement learning~\cite{ng2000algorithms, abbeel2004apprenticeship, ratliff2006maximum, Wulfmeier2015a, finn2016guided} and occupancy matching~\cite{ho2016generative}, we consider behavior cloning~\cite{pomerleau1989alvinn} in this paper due its stability and ease of use. We train a policy $\pi_\theta$, parameterized using a neural network with parameters $\theta$ that takes the observation $\obs_t$ as input and outputs a distribution over actions. The parameters $\theta$ of the policy are trained with stochastic gradient descent to minimize the following loss function:

\begin{equation}
\bcloss(\theta) = \sum_{n=1}^N\sum_{t=1}^T -\log \pi_\theta(\act_t^n |\obs_t^n)
\label{eq:loss_bc}
\end{equation}

\noindent where $\pi_\theta(\act_t |\obs_t)$ represents the distribution over actions for observation $\obs_t$.

\noindent \textbf{One-Shot Imitation Learning.}
In the one-shot imitation learning setting~\cite{duan2017one,finn2017one}, the goal is to learn a meta-policy that can adapt to new, unseen tasks from just a single demonstration for that task. In order to achieve this for tasks with high-dimensional observations such as pixels, which typically require large datasets to learn an effective policy~\cite{vr_imitation, Rahmatizadehmultitaskimitation}, we need to transfer knowledge from demonstrations of previously seen tasks to the task at hand.
Thus, instead of assuming access to expert demonstrations for a single task, one-shot imitation learning assumes an unknown distribution of tasks $p(\task)$, and is provided with a set of tasks $\{\task_i\}$ from this distribution, which are called meta-training tasks.
More concretely, for each training task $\task_i$, we have access to a set of demonstrations $\data_i = \left\{d^i_1, d^i_2,...d^i_N \right\}$. Different tasks contain different objects, and different actions can be performed on those objects. For example, as shown in Figure~\ref{fig:tasks}, a mug could be picked up, a plate could be pushed across a table, a button could be pressed, and a glass could be placed on a tray. The combination of an action and an object constitutes a unique task. 

One-shot imitation learning techniques learn a meta-policy $\pi_\theta$, which takes as input both the current observation $\obs_t$ and a demonstration $\demo$ corresponding to the task which is to be performed, and outputs a distribution over actions. The demonstration $\demo$ specifies to the meta-policy what task is to be performed, and conditioning on different demos can lead to different tasks being performed for the same observation. At training time, we first sample a task $\task_i$, and then sample two demonstrations $d_m$ and $d_n$ corresponding to this task for $m \neq n$. We condition the meta-policy on one of these two demonstrations, say demonstration $\demo_n$, and optimize the following loss on the expert observation-action pairs from the other demonstration, $\demo_m$:

$$
\bcloss(\theta, \demo_m, \demo_n) = \sum_{t=1}^T - \log \pi_\theta(\act_t^m |\obs_t^m, \demo_n) 
$$
We obtain the complete one-shot imitation learning loss by summing across all tasks and all possible demonstration pairs that can be drawn from the same task: 
\begin{equation}
\label{eq:loss_oil}
\loss_\text{oil}(\theta, \fulldataset) = \sum_{i=1}^M \ \sum_{\demo_m,\demo_n \sim \data_i} \bcloss (\theta, \demo_m, \demo_n)
\end{equation}
where $M$ is the total number of training tasks.
\section{Problem Statement}

In this section, we describe our problem setting, and highlight its differences from the standard imitation and reinforcement learning problem settings. Similar to the one-shot imitation learning problem setting described in Section~\ref{sec:prelim}, our goal is to learn a meta-policy that can adapt to new, unseen tasks from just a single demonstration for that task, and we assume access to a set of expert trajectories $\data_i$ for a subset of the meta-training tasks $\{\task_i\}$. 
However, unlike the one-shot imitation learning setting, which only assumes access to a static set of demonstrations, we also assume that we can attempt new trials $\traj$  on the meta-training tasks $\{\task_i\}$ using a meta-policy trained only on $\fulldataset$.

This is similar to the multi-task reinforcement learning and meta-reinforcement learning settings~\cite{karolmultitask, duan2016rl}, but we do not assume access to any task-specific reward functions for any of the tasks. We may also wish to constrain the space of learnable behaviors so that the robot avoids associating ``knocking objects onto the floor'' or ``not touching any objects'' as meaningful tasks. This requires access to a filtering function $\mathcal{F}$, which operates on the trial $\tau$ and returns \texttt{TRUE} if any useful behavior was performed during a trial. This could be automated via a learned classifier for ``useful'' behaviors, or implemented manually via a human annotator that annotates trajectories collected by the meta-policy. Note that the human annotator does not need to specify what task was achieved by the robot in a particular trial: it could be pushing a bowl, picking up a mug or pressing a button, but the human does not need to provide a task label, as long as it is ``useful''. This makes it possible to scale to hundreds of tasks without increasing the difficulty of the annotation. If the filtering function is not provided, the robot can in principle learn to perform all possible behaviors, including undesirable behaviors such as throwing objects onto the floor. While this is not necessarily unreasonable, it could in practice crowd out the desirable behaviors and increase training time.

\section{MILI: Multi-Task Imitation Learning with Improvement}
\label{sec:method}

\begin{algorithm}[tb]
   \caption{\small MILI: Multi-Task Imitation Learning with Improvement}
   \label{alg:metatrain}
\begin{algorithmic}[1]
   \small
   \STATE {\bfseries Input:} Training tasks $\{\task_i\}$, dataset $\fulldataset $
   \STATE {\bfseries Input:} Data collection batch size $B$
   \STATE {\bfseries Input:} Pairing threshold $\alpha$ 
   \STATE \COMMENT{\textit{One-shot imitation pre-training}}
   \STATE $\theta \leftarrow \argmin_\theta \loss(\theta, \fulldataset)$ (see Equation~\ref{eq:final_loss})
   \STATE \COMMENT{\textit{Autonomous data collection and improvement}}
   \STATE Initialize empty trial dataset $\mathcal{R}$
   \FOR{iter$=1,\dots$}
   \FOR{trial$=1,\dots,B$} 
   \STATE Sample training task $\task_i$
   \STATE Sample demo $\demo$ from $\data_i$
   \STATE Collect trial $\tau_n$ from task  $\task_i$ with meta-policy $\pi_\theta(\act|\obs,\demo)$ 
   \IF{$\mathcal{F}(\tau_n)$}
   \STATE Add $\tau_n$ to $\mathcal{R}$
   \ENDIF
   \ENDFOR
   \FOR{ $\tau_n$,$\tau_m$ in $\mathcal{R}$}
   \STATE Compute $p = f_\theta(\tau_n)^T f_\theta(\tau_m)$
   \IF{ $p > \alpha$}
   \STATE Create new task dataset $\data := \{\tau_n, \tau_m\}$
   \STATE $\fulldataset = \fulldataset \bigcup \data$
   \ENDIF
   \ENDFOR
   \STATE $\theta \leftarrow \argmin_\theta \loss_\text{oil}(\theta,\fulldataset)$ (see Equation~\ref{eq:loss_oil})
   \ENDFOR
  \RETURN $\pi_\theta$
\end{algorithmic}
\end{algorithm}

As shown in Figure~\ref{fig:pipeline}, we extend a multi-task, meta-imitation learning pipeline with the ability to learn from the policy's own experience, while preserving the stability and simplicity of supervised learning methods. In order to achieve this, we need to (a) learn a policy that can perform a diverse set of behaviors in a variety of different environments, (b) utilize this policy to autonomously collect data in new environments, and (c) learn an improved one-shot imitation policy that leverages both the initial dataset and the autonomously collected data. For obtaining a data collection policy, we perform meta-imitation learning on a human-provided dataset. We condition this meta-policy on random demos for collecting trials in new environments. We also learn a \textit{latent space} of skills using the same initial dataset, which is then used for organizing the collected trials into new tasks. We now run meta-imitation learning on this expanded dataset, resulting in an improved policy that utilizes both human-provided and autonomously collected data.

\noindent \textbf{Learning a data collection policy} Na\"{i}vely performing behavior cloning (as described in Section~\ref{sec:prelim}) on the human-provided dataset is unlikely to result in diverse behavior: since a robot can perform many useful tasks in any given scene, cloning the actions from all human-provided trajectories without providing any context will lead to an averaging of the different possible actions, and the behavior generated by the policy will not be useful. Therefore, we instead train a demo-conditioned meta-policy using the loss function described in Equation~\ref{eq:loss_oil} (see Section~\ref{sec:prelim}). We can now use this meta-policy to collect trials in new scenes by conditioning it on different demonstrations from $\fulldataset$. 

\noindent \textbf{Utilizing autonomously collected data} A straightforward way to improve our meta-policy's performance using the collected trials would be to optimize the one-shot imitation learning loss defined in Equation~\ref{eq:loss_oil} on the collected trajectories. However, in order to optimize this loss, we need at least two successful trials for any given task, as described in Section~\ref{sec:prelim}. That is, we need at least two trials depicting optimal behavior for the same task for it to be useful for learning. Since our filtering function only labels trials as being useful for any task, we cannot use it for assigning trajectories to specific tasks. If we can find two trials that perform a similar behavior, we can add them to our dataset as corresponding to a new task. 

However, finding similar trials can be non-trivial: we wish to learn policies from high-dimensional observations spaces such as visual inputs, which implies that our trials consist of videos, and finding the distance between two videos by computing a standard distance metric like the Euclidean distance is unlikely to result in useful pairings. This motivates the need to learn a \textit{latent space of tasks}, in which we can embed any given trial, and compute meaningful distances between the trials. If two trials are found to be close to each other in the latent space, we can add them to $\fulldataset$ as a pair of demonstrations corresponding to a new task, and update the meta-policy to minimize the loss in Equation~\ref{eq:loss_oil}. In Section~\ref{sec:latent_space}, we describe how we can learn a latent space of tasks jointly with the meta-policy, using only the human-provided one-shot imitation dataset and no extra supervision. In Section~\ref{sec:improvement}, we describe how we utilize a learned latent space of tasks and meta-policy to improve from autonomously collected data.

\subsection{Learning a Latent Task Space Jointly with the Policy}
\label{sec:latent_space}
\begin{figure}
\centering
  \includegraphics[width=1.0\columnwidth]{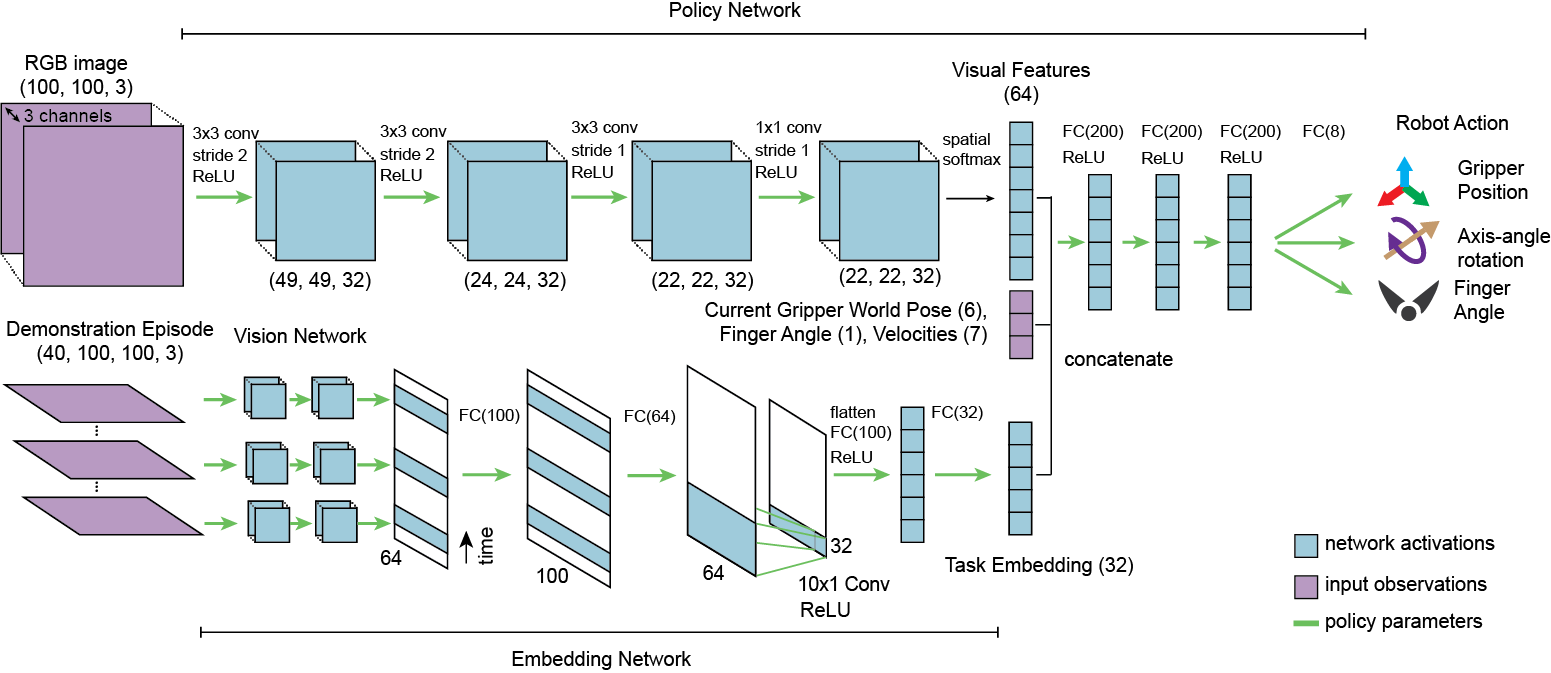}
  \vspace{-0.6cm}
  \caption{We train visuomotor meta-imitation policies end-to-end, based on the setup described by \cite{wtl}. Upper left and right: RGB observations are mapped to visual features, which are transformed into desired gripper positions. Lower: The task embedding network aggregates visual features over a sequence of image observations (obtained from a demonstration episode or policy rollout) into a ``task embedding'', which is used to condition multi-task behavior from the policy network. Unlike \cite{wtl}, the task embedding network does not condition on observed gripper positions in the demo/rollouts.}
  \vspace{-0.6cm}
  \label{fig:arch_diagram}
\end{figure}

Following prior work in one-shot imitation learning~\cite{duan2017one, james2018task, wtl}, we learn a meta-policy $\pi_\theta(\act | \obs, \demo)$ that consists of two neural networks: an embedding network $f_\theta$ and a policy network $g_\theta$, where $\theta$ represents the parameters of the two neural networks (shown together in Figure~\ref{fig:arch_diagram}). Below, we detail the roles performed by these networks, their architecture and loss functions that we use to train them.

\noindent \textbf{Embedding network and contrastive loss functions.}
The embedding network, represented as $f_\theta$, accepts as input a demo of the task to be performed. The embedding network consists of a convolutional neural network followed by 1-D temporal convolutions, and embeds the demo into a fixed-length vector, which we denote as $f_\theta(d)$, and refer to as the demo embedding. Intuitively, we would like the embeddings to satisfy the following property: we want two demo embeddings to be close to each other if the demos correspond to the same task, and we wish them to be further apart for demos that belong to different tasks. Formally, this can be accomplished by considering the distance function $H_\theta(d_m,d_n) = \|f_\theta(d_m) - f_\theta(d_n)\|_2^2$. $H_\theta$ should be low when $d_m$ and $d_n$ correspond to the same task, and it should be high when they correspond to different tasks. Contrastive loss functions~\cite{Hadsell06} satisfy this property, and are given by the following expression: 
\begin{align*}
\loss_\text{c}(\theta,\fulldataset) = \sum_{j=1}^M \sum_{k=1}^M  \sum_{\substack{\demo^j_m \sim \data_j\\ \demo^k_n \sim \data_k}} & \mathbbm{1}(j=k)H_\theta(\demo^j_m,\demo^k_n) +\\
& \mathbbm{1}(j\neq k) \max(0,\beta - H_\theta(\demo^j_m,\demo^k_n))),
\end{align*}
where $\beta$ is a margin (which we set to 1.0).

\noindent \textbf{Policy network.}
The remaining part of the meta-policy consists of a policy network. It takes as input an image of the current scene (along with other parts of the robot's state such as end-effector pose), and outputs a distribution over actions. In order to learn to predict actions from this information, we use the one-shot imitation learning loss $\loss_\text{oil}$  in Equation~\ref{eq:loss_oil} in Section~\ref{sec:prelim}.
Our complete loss function,
\begin{equation}
\loss(\theta,\fulldataset) = \loss_\text{oil}(\theta,\fulldataset) + \loss_\text{c}(\theta,\fulldataset)
\label{eq:final_loss}
\end{equation}
is minimized using the Adam~\cite{adam} optimizer, and we train both the embedding and policy networks jointly, sharing the convolutional layers between them. The output of this procedure results in the meta-policy $\pi_\theta$, where $\theta$ denotes the learned parameters.

\subsection{Autonomous Improvement}
\label{sec:improvement}
Our method is summarized in Alg.~\ref{alg:metatrain}. We utilize the learned meta-policy $\pi_\theta$ for collecting data in new scenes. Given a scene, we condition the learned meta-policy on a randomly sampled demo from the initial dataset $\fulldataset$ (see Lines 9-11 in Alg.~\ref{alg:metatrain}). We then run the meta-policy in this new scene, and add it to the trial dataset $\mathcal{R}$ if the filtering function $\mathcal{F}$ deems the trial to have performed some useful task (but does not provide any label about what the performed task was). We then search the trial dataset $\mathcal{R}$ for similar trials by computing the cosine distance (i.e. a normalized dot-product) between the embedding of the newly collected trial, and all trials collected in the past (see Lines 17-18 in Alg.~\ref{alg:metatrain}). If the cosine distance for any pair of trials is found to be above a pre-specified threshold $\alpha$ (which we determine using cross-validation to be 0.9), we add the pair of trials to the dataset as corresponding to a new task (see Lines 19-21 in Alg.~\ref{alg:metatrain}). We then update our meta-policy $\pi_\theta$ using the one-shot imitation learning loss (see Line 24 in Alg.~\ref{alg:metatrain}).

\section{Experiments}
\noindent Our experiments seek to answer the following questions:
\begin{enumerate}
\item Does our method enable autonomous improvement? That is, can the trials generated by the policy in new environments improve one-shot imitation performance in comparison to only meta-imitation learning or only behavior cloning on a static dataset?
\item How does the performance vary with the number of trials collected?
\item How does the performance of our learned latent space-based pairing model compare to an oracle that pairs the autonomously collected data optimally?
\end{enumerate}

\begin{figure}[t]
\centering
  \includegraphics[width=1.0\columnwidth]{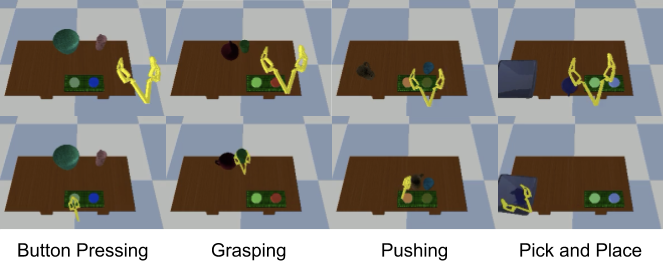}
  \vspace{-0.8cm}
  \caption{Our dataset of tasks consists of four distinct task families: button pressing, grasping, pushing and pick and place. Within each task family, we have hundreds of tasks, and our dataset contains over a hundred distinct kitchenware objects. Our train set has 520 tasks, while our validation and test sets each have 40 tasks each. We tune hyperparameters on the validation tasks, and report final performance on the test tasks.}
  \vspace{-0.6cm}
  \label{fig:tasks}
\end{figure}

We conduct our experiments on a realistic 3D simulation created using the Bullet physics engine~\cite{coumans2016pybullet}, shown in Figure~\ref{fig:tasks}. It consists of a 7-DoF robotic gripper controlled with continuous position control from visual observations at 10Hz to accomplish manipulation tasks from four distinct task families, and contains over a hundred different kitchenware objects~\cite{vrgripper}. 
The four task families that we consider in our experiments are: button pressing, grasping, pushing and pick and place. Example tasks from each of the four families are shown in Figure~\ref{fig:tasks}.
Different tasks within these families correspond to different sets of objects and manipulating those objects in different manners.
For example, a pushing task in a scene containing both a mug and a bowl could be to push the bowl to the mug, or it could be to push the mug to the bowl.
To succeed at a task, the robot must be able to perform the task from multiple different initial object arrangements such that it cannot simply memorize the motion from the one provided demonstration.
The policy network, visualized in Figure~\ref{fig:arch_diagram}, uses pixel inputs for all the tasks, i.e. an RGB image of size 100$\times$100, which gets passed in to the control network along with the 7-DoF gripper pose, three dimensions of which correspond to the its position in 3D space, three correspond to the Euler angles, and one corresponds to the finger angle. The output of the policy is the desired 7-DoF pose of the gripper, represented as the 3D position. We train on 520 tasks, use a validation set of 40 tasks for hyperparameter tuning, and show performance on a test set of 40 tasks, which contain held-out objects.

Demonstrations for the initial dataset $\fulldataset$ are collected by a human using an HTC Vive virtual reality system. Four demonstrations are collected for each task, and the object positions are randomized between demonstrations.
At evaluation time, we are provided with one demonstration for a given test task, and the policy performance is evaluated with an object arrangement that is different from the demonstration. 

\begin{figure}[h]
\centering
  \includegraphics[width=1.0\columnwidth]{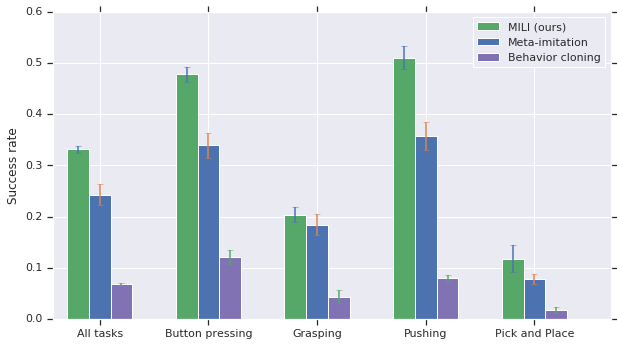}
  \vspace{-0.8cm}
  \caption{We evaluate one-shot imitation performance of our policy on 40 unseen test tasks from four distinct task families. We ran our method, as well as both of the comparisons, with five random seeds, and report the average final performance and standard error across seeds. All methods use the same human demonstration data. We see that our method outperforms the meta-imitation learning baseline for all task families, indicating that the autonomously collected experience results in a better policy.}
  \vspace{-0.4cm}
  \label{fig:results_data}
\end{figure}

\subsection{Autonomous Improvement}
We first evaluate the central question of our paper: can the robot collect trials in new scenes and use them to improve its performance? We bootstrap the meta-policy from 800 human teleoperation demos on 200 different tasks, and learn a one-shot imitation meta-policy as well a latent task space from this data (see Line 5 in Algorithm~\ref{alg:metatrain}). We collect 60K trials from 150 different scenes. We find that 8.2K of these trials pass the filtering function $\mathcal{F}$ and get added to our trial dataset $\mathcal{R}$. Our method finds pairs of similar trials using a pairing threshold of $\alpha=0.9$ in the trial dataset
$\mathcal{R}$, adds them to $\fulldataset$, and re-runs meta-imitation learning (see Lines 17-24 in Algorithm~\ref{alg:metatrain}). The performance of this policy is shown in Figure~\ref{fig:results_data}, and videos can be found on our project
website\footnote{Project page: \url{https://sites.google.com/view/scalable-mili/}}. While it is possible to run several iterations of our method for continuous improvement, we only evaluate one round of our method in this section. We compare our method against:

\noindent \textbf{Meta-imitation} This corresponds to optimizing the loss function described in Equation~\ref{eq:loss_oil} on the human demonstrations dataset $\fulldataset$, and is representative of prior work~\cite{james2018task}.

\noindent \textbf{Behavior cloning} This corresponds to optimizing the loss function described in Equation~\ref{eq:loss_bc} on all the demos from $\fulldataset$. 

As shown in Figure~\ref{fig:results_data}, our method substantially outperforms the meta-imitation learning baseline, achieving a relative improvement of 36.8\%. The meta-imitation baseline in turn outperforms the behavior cloning baseline, which is expected since the behavior cloning baseline does not incorporate task-specific information. This shows that our proposed method is capable of incorporating policy roll-outs in the learning process so as to improve the policy's one-shot imitation performance on new, unseen tasks. 

\subsection{Varying the Number of Trials}
A central claim in our paper is that a policy can continuously improve if it is able to collect more and more data and use it for learning. In this section, we aim to answer the following question: do we achieve monotonically improving performance as we collect more trials and add them to the dataset? To answer this question, we run our method with five random seeds while varying the number of trials collected: 10K, 20K, 40K and 60K. The results of this experiment are summarized in Figure~\ref{fig:results_data}, and show the following trend: as we keep collecting more trials, the one-shot imitation performance of our meta-policy for unseen tasks keeps improving. The relative improvements in performance are highest for the first 10K trials collected, but the performance starts levelling off as we collect more data.

\subsection{Using Oracle Task Labels}
Evaluating meta-imitation learning on an optimally paired dataset of policy rollouts allows us to measure what success rates we might hope MILI to achieve if we learned a perfect latent space. This oracle achieves a one-shot imitation success rate of $33.7\%$ as compared to MILI's 32.7\%, suggesting that MILI can drastically reduce the burden of labeling tasks without significant loss in performance. 

\begin{figure}[t]
\centering
  \includegraphics[width=0.85\columnwidth]{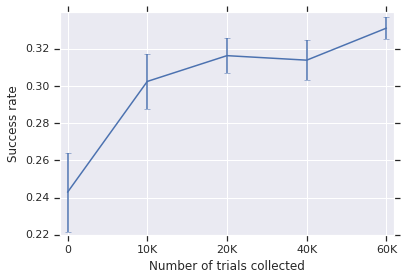}
  \caption{Performance of MILI improves as we increase the number of trials.
  }
  \vspace{-0.4cm}
  \label{fig:results_numtrials}
\end{figure}
\section{Discussion and Future Work}

In this paper, we study how meta-imitation learning can be used to enable an agent to improve with autonomously collected data. Our aim is to preserve all of the desirable properties of imitation learning -- simplicity and stability -- while augmenting our method with one of the most powerful features of RL: the ability to improve from autonomously collected trials. Our approach is based on a simple observation: in a multi-task setting, the robot's attempts to perform a new task, even if unsuccessful, may still serve as successful trials for \emph{other} tasks. Therefore, if we can match them with corresponding prior trials and thereby form new task datasets for meta-learning, we can incorporate the robot's own attempts as pseudo-demonstrations for other tasks. Unlike RL, our method does not require any task-specific reward function, only a general, task-agnostic filtering function that indicates whether a given trial is useful for \emph{any} task. This makes it straightforward to scale our approach to large skill repertoires.
After bootstrapping our meta-imitation policy from a few hundred human-provided demonstrations, we can collect tens of thousands of autonomous trials, and use them to meta-train a final policy that substantially outperforms standard meta-imitation learning on the initial dataset. 

\noindent \textbf{Limitations.} Our work opens several possible directions for future work. Currently, our method still requires a filtering function, which means we need to obtain a binary label from a human oracle for each trial that we collect to decide if it should be included in the training set. While this signal is easy to obtain, future work might investigate learning-based methods that perform this filtering automatically, further boosting the scalability of this approach. 
Further, while our method improves with more data, the overall success rates are still relatively low due to the challenging nature of the environment. We expect that further advances in vision-based meta-learning will complement the approach in this paper to boost performance.
Finally, another exciting direction for future work is to devise a complete lifelong learning approach based on our method, performing repeated iterations of data collection and meta-imitation on a real-world robotic system so as to continually improve the policy on both the original set of tasks and new tasks discovered during data collection.

\end{document}